\documentclass[10pt,twocolumn,letterpaper]{article}

\usepackage{iccv}
\usepackage{times}
\usepackage{epsfig}
\usepackage{graphicx}
\usepackage{amsmath}
\usepackage{amssymb}
\usepackage{caption}
\usepackage{subcaption}
\usepackage{booktabs}
\usepackage[thinc]{esdiff}
\usepackage{algorithm}
\usepackage{algpseudocode}

\usepackage[pagebackref=true,breaklinks=true,letterpaper=true,colorlinks,bookmarks=false]{hyperref}
\usepackage{authblk}

\iccvfinalcopy 

\def\iccvPaperID{32} 

\ificcvfinal\fi

\begin{document}
\author[1]{Zikun Chen}
\author[2]{Han Zhao}
\author[3]{Parham Aarabi}
\author[1]{Ruowei Jiang}

\affil[1]{ModiFace Inc.}
\affil[2]{University of Illinois at Urbana-Champaign}
\affil[3]{University of Toronto}
\renewcommand*{\Authands}{, }

\title{SC$^2$GAN: Rethinking Entanglement by Self-correcting Correlated GAN Space}

\maketitle

 \ificcvfinal\thispagestyle{empty}\fi


\begin{abstract}

    Generative Adversarial Networks (GANs) can synthesize realistic images, with the learned latent space shown to encode rich semantic information with various interpretable directions. However, due to the unstructured nature of the learned latent space, it inherits the bias from the training data where specific groups of visual attributes that are not causally related tend to appear together, a phenomenon also known as spurious correlations, e.g., age and eyeglasses or women and lipsticks. Consequently, the learned distribution often lacks the proper modelling of the missing examples. The interpolation following editing directions for one attribute could result in entangled changes with other attributes. To address this problem, previous works typically adjust the learned directions to minimize the changes in other attributes, yet they still fail on strongly correlated features. In this work, we study the entanglement issue in both the training data and the learned latent space for the StyleGAN2-FFHQ model. We propose a novel framework SC$^2$GAN that achieves disentanglement by re-projecting low-density latent code samples in the original latent space and correcting the editing directions based on both the high-density and low-density regions. By leveraging the original meaningful directions and semantic region-specific layers, our framework interpolates the original latent codes to generate images with attribute combination that appears infrequently, then inverts these samples back to the original latent space. We apply our framework to pre-existing methods that learn meaningful latent directions and showcase its strong capability to disentangle the attributes with small amounts of low-density region samples added.
\end{abstract}

\section{Introduction}

%

Recent advances of Generative Adversarial Networks (GANs)~\cite{goodfellow2014generative}, such as StyleGAN~\cite{karras2019style,karras2020analyzing,karras2021alias} and BigGAN \cite{brock2018large}, boost a remarkable success for synthesizing photo-realistic images.
In addition to a variety of real-world applications such as image-to-image translation \cite{Isola2017CVPR,zhu2017unpaired} or text-to-image translations \cite{zhang2017stackgan,li2019controllable}, another line of work \cite{bau2019gandissect,bau2019seeing,wu2021stylespace,harkonen2020ganspace,shen2020interfacegan,chen2022exploring} that studies the interpretability of GANs has also generated increasing attention in the community.
These works study the learned latent space by identifying semantically meaningful directions and interpolating along the learned directions. However, challenges remain to perfectly disentangle correlated features such as age and eyeglasses while obtaining valid feature controls, due to the biased learned distribution.
\begin{figure}[t]
\hspace{-0.3cm}
\includegraphics[width=1\linewidth]{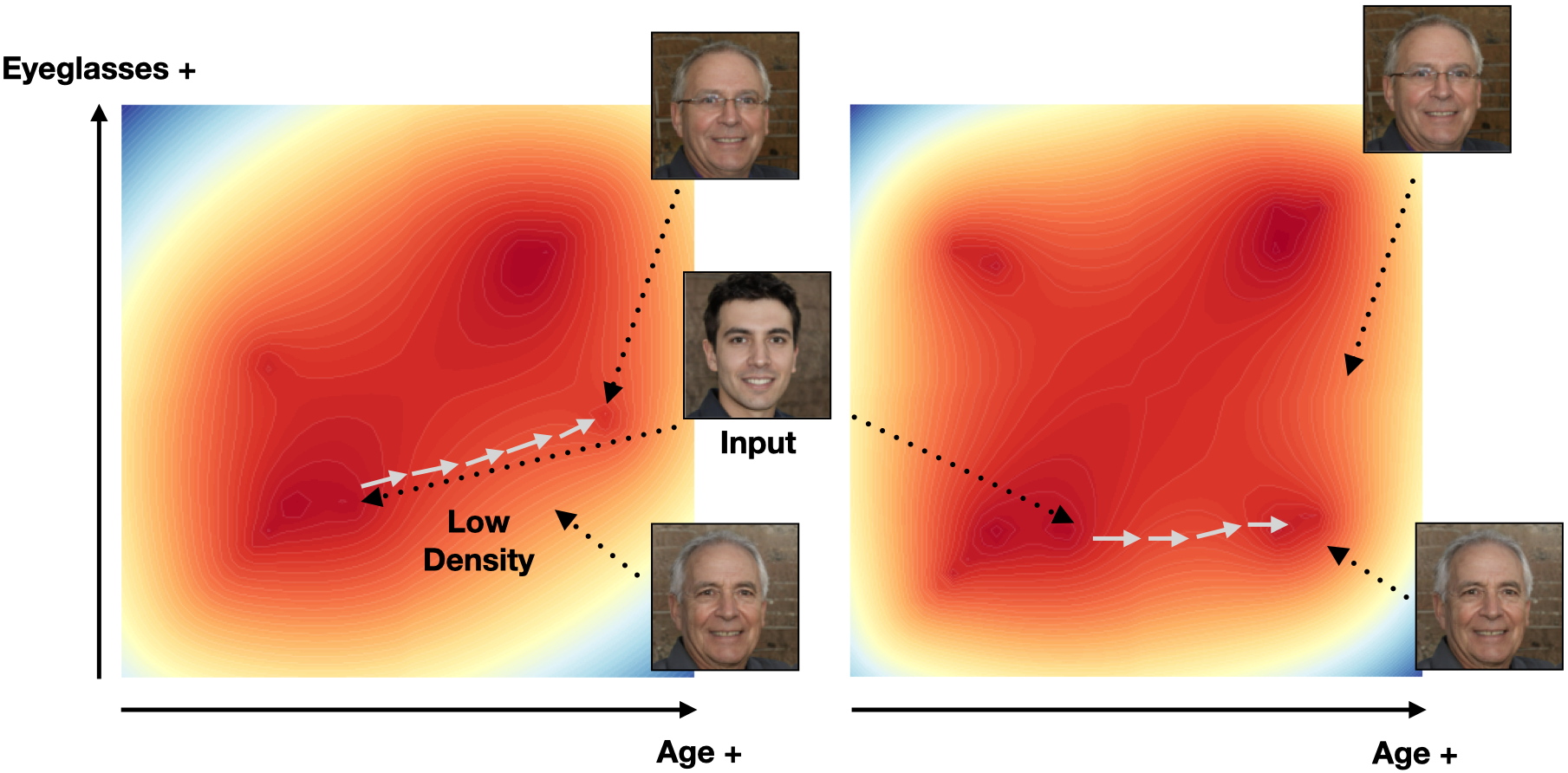}
   \caption{Illustration of our method: original latent distribution (left) and self-corrected latent distribution (right). The intensity of the color indicates the density of different sub-regions in the learned latent space. The white arrows show the interpolation directions. The ``eyeglasses'' direction and the ``age'' direction are more orthogonal to each other in the self-corrected latent space. }
\label{fig:overview}
\vspace{-0.1cm}

\end{figure}

To tackle the challenge of entangled features, prior works have largely taken three approaches to obtain disentangled controls: orthogonalization of the learned directions \cite{shen2020interfacegan,wang2022provable,balasubramaniam2022invariant}, controls based on semantic masks \cite{wu2021stylespace}, and gradient-based channel filtering \cite{chen2022exploring}. Orthogonalization of the learned directions~\cite{shen2020interfacegan,wang2022provable,balasubramaniam2022invariant} follows the assumption that for any learned direction, a change along another orthogonal axis should not affect the feature for the learned direction. For example, if $A$ and $B$ are two orthogonal directions that define two hyperplanes, changing along the $A$ axis should not affect its distance to $B$'s hyperplane. This can be achieved through projecting one direction onto another or optimization-based procedures. In practice, as shown in \cite{shen2020interfacegan}, the learned directions are often found as orthogonal yet entangled in the embedding vector space because projection can only remove linear correlations while nonlinear relationships between them could still exist. The second type of strategy~\cite{wu2021stylespace} utilizes the information within the semantic mask and disentangles the features in different semantic regions. While this shows its effectiveness over more localized attributes, it fails to generate controls for global attributes like gender or age. Gradient-based channel filtering~\cite{chen2022exploring} selects channels based on the importance with respect to a target attribute. More concretely, by taking the gradient with respect to each attribute, Chen et al.~\cite{chen2022exploring} select the channels that have the maximal impact on the target attribute while filtering out the channels with the maximal impact on other attributes. However, this could fail if two attributes are strongly correlated and share almost the same set of channels for decisions.

What if the original GAN space is ``entangled"? In this case, will the GAN model be able to generate images that never or hardly appear in the training data? For instance, images which are typically out of the distribution of the training data such as men with lipstick or women with beards. We hypothesize that the lack of such training data results in non-uniformly distributed density (hence entanglement) in the learned latent space, which leads to the bias of the identified directions. To support our intuition, we first show the empirical findings of the correlation between different attribute pairs in the original image distribution and how this affects the learned GAN space.
Inspired by our empirical findings, we propose a novel framework called SC$^2$GAN to obtain disentangled controls. In particular, we project generated samples with GAN inversion methods back into the low-density regions in the learned latent space to achieve a more balanced latent space distribution, which can help decorrelate the pairs of attributes. We show that the interpretable directions re-learned by different methods under our framework would be corrected towards the correct cluster, as shown in the right panel of \textbf{Figure~\ref{fig:overview}}. 

Our contributions can be summarized as follows:
\begin{itemize}
    \item We study the entanglement problem by showing the unbalancedness of certain attribute pairs in the original dataset and the resulting bias in the learned latent space.
    \item We propose a simple yet effective framework called SC$^2$GAN to disentangle features by correcting the biased latent space via projecting certain manipulated sample groups.
    \item We qualitatively and quantitatively show the effectiveness of our method, which is applicable as a post-processing procedure to many existing approaches that learn interpretable directions.
\end{itemize}

\section{Related Work}
We provide an overview of two different categories of approaches to control GAN outputs, as well as the line of work that embeds real images into GAN latent space.
\subsection{Image Editing with Conditional GANs}
By incorporating class label-related loss terms during training, conditional GANs obtain explicit controls over the generation process~\cite{mirza2014conditional,Isola2017CVPR,odena2017conditional}, which can generate images of classes specified by the user with a class label as input. Nevertheless, they lack controls over fine-grained attributes hence the entanglement issue can still occur. Recently, in the face image generation domain, new methods have been proposed to gain more fine-grained controls over multiple attributes~\cite{deng2020disentangled,tewari2020stylerig}. These approaches translate 3D face rendering controls, i.e., 3DMM~\cite{blanz1999morphable} parameters, into the GAN framework, and are able to control the expressions, pose and illumination while preserving the identity. However, controls learnable by these methods are limited to existing 3D models' parametrization of facial attributes. 
 \subsection{Interpolation in GAN Latent Space}
Unlike conditional GANs, another line of work~\cite{bau2019gandissect,shen2020interfacegan,harkonen2020ganspace,wu2021stylespace} explores controls over output image semantics in GANs trained without labels. They have shown that such GAN latent space encodes rich semantic information with numerous meaningful directions, interpolation along which results in human-interpretable changes in the output semantics.
InterFaceGAN~\cite{shen2020interfacegan} employs pre-trained image classifiers to cluster latent codes corresponding to different semantics and trains SVMs on those samples to learn the editing direction.
Grad-Control~\cite{chen2022exploring} works similarly by training fully connected layers on a small amount of labelled latent codes, and taking the classifier gradient directions as the meaningful path in the latent space.
GANSapce~\cite{harkonen2020ganspace} works in an unsupervised way by performing PCA on features in the generator, and regressing linear directions in the latent space corresponding to the principal components, which correspond to human-interpretable changes in the image space.
StyleSpace~\cite{wu2021stylespace} learns more fine-grained controls by computing latent channels exclusively activated for semantic regions defined by pre-trained semantic segmentation networks. 

Although various semantic directions have been discovered, during interpolation, entanglement in attributes, i.e., changing the target affects other causally independent attributes, often occurs. This phenomenon is known as spurious correlation~\cite{wang2022provable,balasubramaniam2022invariant} and could be ascribed to the nature of the learned latent space, i.e., groups of visual attributes are not guaranteed to be uniformly distributed in the training data, hence the generator captures such spurious correlations and implicitly encodes them in its latent space. 
To address this issue, in the context of GAN latent space, Shen et al.~\cite{shen2020interfacegan} proposes to adjust the editing directions and minimize the change in the entangled attributes by orthogonalizing the target direction from the entangled attribute through projection, while~\cite{chen2022exploring} filters out salient latent channels for predicting the entangled attribute during interpolation. More generally, if the structure of the causal graph is known, Wang et al.~\cite{wang2022provable} propose a post-processing method to provably identify and filter the feature subspace spanned by these spurious features by projecting the features to the so-called invariant feature subspace.

Although the aforementioned methods achieve partial success, disentanglement remains challenging when the correlation between attributes is significantly strong. After the adjustment, the resulting direction often only brings trivial changes in the target, or very few channels are left for the target attribute.~\cite{wu2021stylespace} suffers less from the entanglement issue as it focuses on attributes belonging to localized semantic regions, but it lacks the ability to edit global attributes like age that require changes over the entire image. Different from approaches that directly adjust the biased directions, we propose to utilize such directions and debias the learned latent space by generating samples in low-density regions, and re-learn the editing directions based on the corrected latent distribution. 

\subsection{GAN inversion}
GAN inversion embeds real-world images into the GAN latent space, which can then be edited with latent space interpolation~\cite{abdal2019image2stylegan,abdal2020image2stylegan++,karras2020analyzing,tov2021designing,wang2022high}. There are two main categories of GAN inversion: optimization-based methods~\cite{abdal2019image2stylegan,karras2020analyzing,abdal2020image2stylegan++}, which sample from the original latent space and optimize the latent code to match the output with the real image target, or encoder-based methods~\cite{tov2021designing,wang2022high}, which aim to invert the generation process and learn the reverse mapping from image space to the latent space, with the help from training on a large number of latent code-image pairs. One common challenge for GAN inversion is the tradeoff between distortion (i.e., resemblance to the target) and editability (i.e., how close the inverted code lies to the original latent distribution for the latent interpolation directions to be applicable), and different regularization methods have been proposed to handle such tradeoffs~\cite{zhu2020improved,tov2021designing}. In our work, we employ latent optimization to obtain samples with infrequent combinations of attributes, as we find it achieves little distortion and the results faithfully represent the minority distributions in the biased learned space. 

\section{Methodology}
In this section, we describe the motivation and details of our method to self-correct StyleGAN $\mathbf{W}$ space and obtain disentangled image manipulations.
We first show our empirical observations of entanglement in the learned latent distribution, followed by a more in-depth analysis that quantifies such phenomena. 
We then demonstrate our proposed novel framework that addressed the disentanglement by balancing the latent distribution. Specifically, we generate images that correspond to the low-density regions and rebalance the distribution by projecting those images back into the learned latent space.

\begin{figure}[b]
\begin{subfigure}{1\linewidth}
\centering

\includegraphics[width=\linewidth]{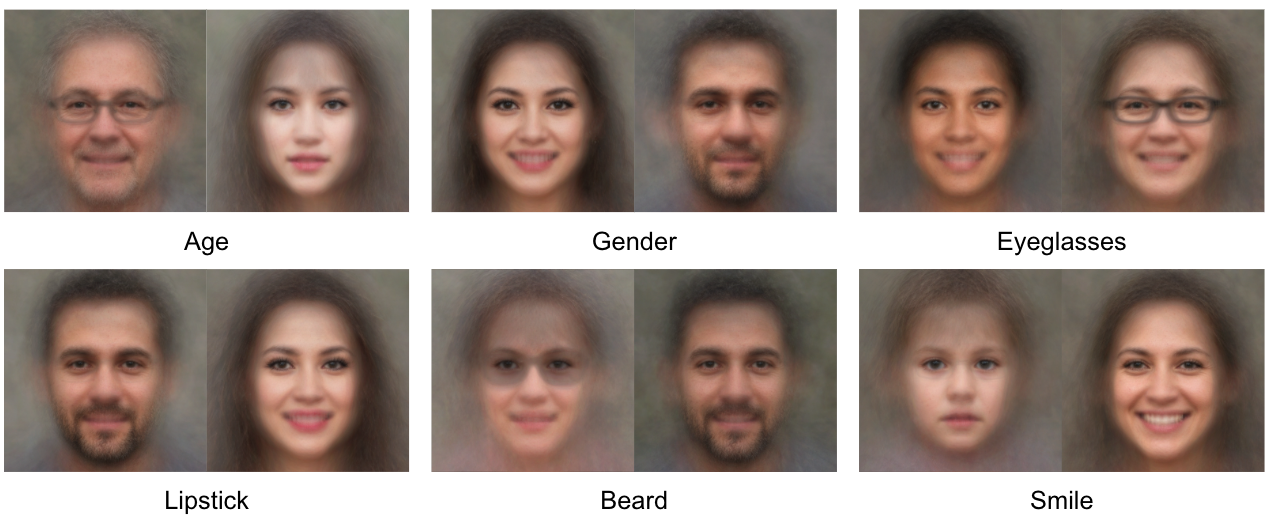}
   \caption{Averaged faces for each attribute sign showing entanglement from the GAN-generated images.}
\label{fig:before_aggregated}

\end{subfigure}
\begin{subfigure}{1\linewidth}
\centering

\includegraphics[width=\linewidth]{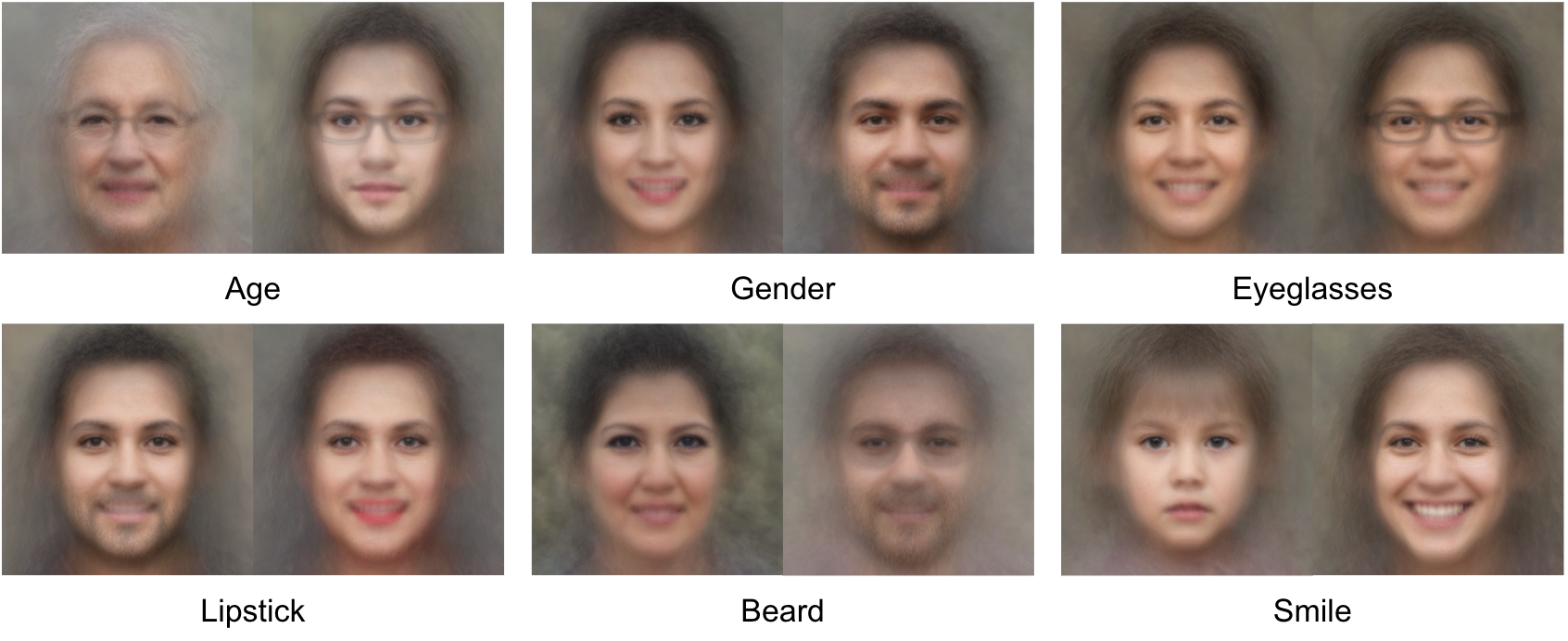}
   \caption{Averaged faces from the merged dataset containing equal amounts of original $\mathbf{W}$ samples and self-corrected samples. By projecting data onto low-density regions in the original clusters, e.g., edited images of old people not wearing eyeglasses or men with lipstick, the corrected distributions show less attribute correlations.}
\label{fig:after_aggregated}

\end{subfigure}

   \caption{Averaged faces for each attribute sign sampled from the original $\mathbf{W}$ space and inverted from $\mathbf{W+}$ edits.}
\label{fig:aggregated}
\end{figure}

\begin{figure*}[t]
\centering
\begin{subfigure}[b]{0.25\linewidth}
    \includegraphics[width=\linewidth]{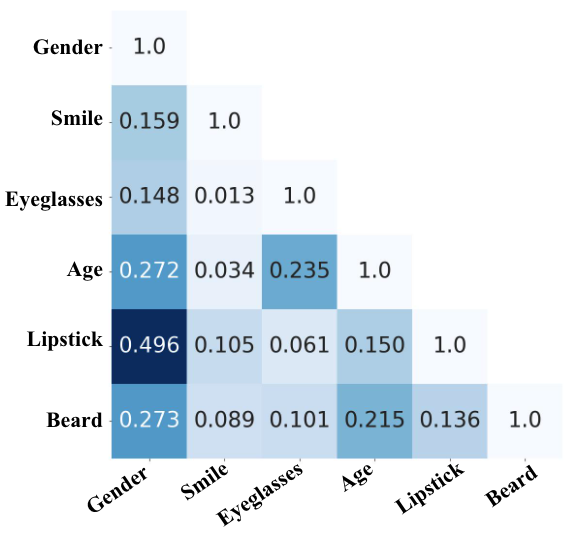}
    \caption{Original FFHQ dataset. }
    \label{fig:ffhq_gandata_corr:ffhq}
\end{subfigure}\hfill
\begin{subfigure}[b]{0.25\linewidth}
    \includegraphics[width=\linewidth]{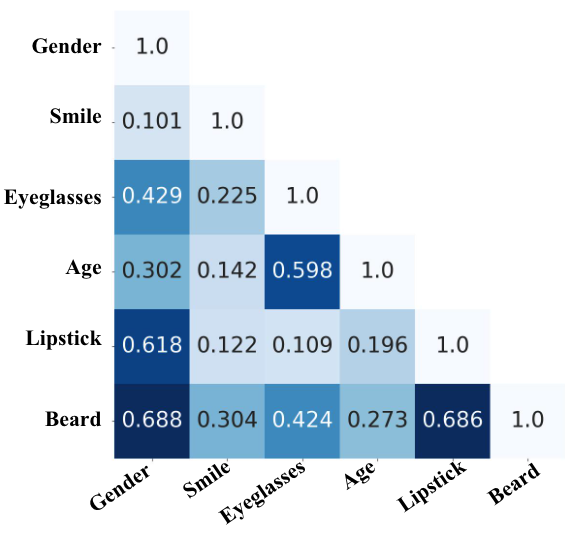}
    \caption{StyleGAN2 generated images.}
    \label{fig:ffhq_gandata_corr:gandata}
\end{subfigure}\hfill
\begin{subfigure}[b]{0.25\linewidth}
    \includegraphics[width=\linewidth]{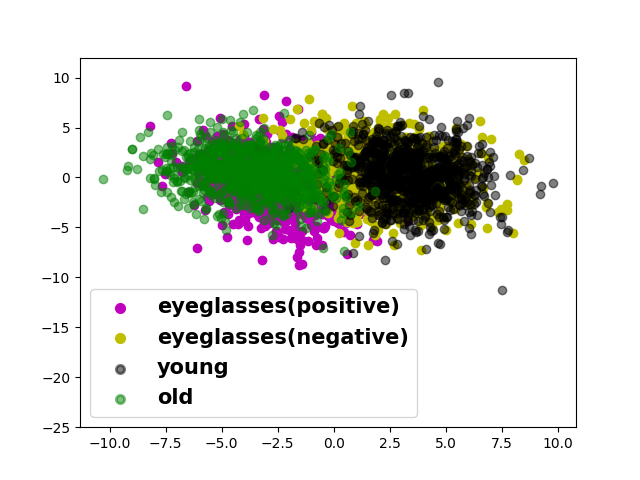}
    \caption{PCA of original.}
    \label{fig:pca_before}
\end{subfigure}\hfill
\begin{subfigure}[b]{0.25\linewidth}
    \includegraphics[width=\linewidth]{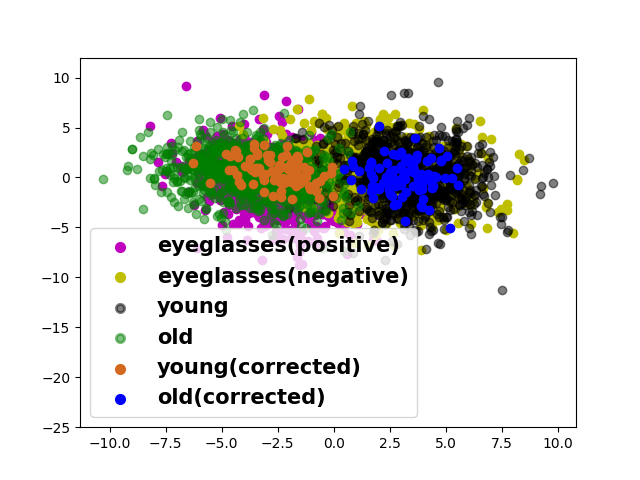}
    \caption{PCA with corrected.}
    \label{fig:pca}
\end{subfigure}
\caption{Illustration of the entanglement. (a) and (b) are the absolute values of tetrachoric correlations between each attribute pair in the original FFHQ dataset and StyleGAN-generated images. (c) is an example visualization of the latent space entanglement between eyeglasses and age, where the decision boundaries and interpolation directions for both attributes are likely to be similar. We show age clusters and eyeglasses clusters, fitted with PCAs trained on eyeglasses samples and projected with the top two components. In (d), we project the self-corrected samples (young with eyeglasses and old without eyeglasses) onto the same axes. }
\vspace{-0.1cm}

\end{figure*}

\subsection{StyleGAN Latent Space Entanglement}
\noindent \textbf{StyleGAN Latent Space.} Our method to rebalance and disentangle the StyleGAN $\mathbf{W}$ space is motivated by observations of correlations between pairs of attributes in StyleGAN outputs. 
GANs map latent codes $z$ from a known distribution $Z \subseteq \mathbb{R}^d$ to an image space $X \subseteq \mathbb{R}^{H\times W\times3}$ with the mapping function~$g : Z \rightarrow X$. 
In StyleGAN~\cite{karras2019style}, instead of directly feeding $z$ to the generative blocks, the output is controlled by a function of $\mathbf{w} = M(z)$ where $M$ is a multilayer perceptron network with 8 layers. 
The vector $\mathbf{w}$ is a style vector lying in the $\mathbf{W} \subseteq \mathbb{R}^d$ space and each $\mathbf{w}$ vector is repeated 18 times to a $\mathbf{w+}$ vector and fed to generator layers at different resolutions to generate the final image with $G(\mathbf{w+} ; \theta))$, which has been shown to enable powerful controls of features at different abstraction levels. 

\noindent \textbf{Biases in learned $\mathbf{W}$ space.} Multiple works~\cite{karras2019style,shen2020interfacegan,harkonen2020ganspace} have discovered that, unlike the original $Z$ distribution, the $\mathbf{W}$ space distribution is distorted as it captures the spurious correlations between attributes in the training data, resulting in low-density regions for the minority attribute groups. 
To visualize such effects, we randomly sampled 500k images from StyleGAN2-FFHQ~\cite{karras2020analyzing} and employed pre-trained CelebA claasifiers~\cite{karras2019style} on our image bank to select the most confident 1k samples for a set of attributes. As shown in \textbf{Figure}~\ref{fig:before_aggregated}, we aggregate the 1k faces by computing the pixel space averages and show strong correlations among different attribute pairs, e.g., men with lipstick and women with beard are often underrepresented in the learned space, which directly relates to the entanglement problem many previous works $\mathbf{W}$~\cite{shen2020interfacegan,chen2022exploring,harkonen2020ganspace} suffer from, where editing one attribute affects the correlated attribute as well.

In order to further explain the findings above, we analyze the latent space entanglement from both the training data and the learned latent distribution perspectives. 
First, we analyze the original FFHQ training data for StyleGAN. 
With ffhq-features-dataset~\cite{mrmartin2019}, \textbf{Figure}~\ref{fig:ffhq_gandata_corr:ffhq} measures the correlations between each pair of attributes, which exhibits non-trivial correlations between attributes like eyeglasses and age in alignment with our observations above. 
Next, we analyze the $\mathbf{W}$ space leveraging knowledge from pre-trained image classifiers~\cite{karras2019style}. 
In particular, with our 500k image bank and pseudo labels for each attribute of interest, we computed the same correlation matrix in \textbf{Figure}~\ref{fig:ffhq_gandata_corr:gandata}.
The big correlations between certain attribute pairs make the learning of disentangled editing directions challenging. 
For instance, for eyeglasses and age, since the high-density region for the old mostly contains old people wearing eyeglasses code samples,
it's highly likely that when interpolating young latent code without eyeglasses following the old direction, eyeglasses will be added. 
Essentially, this corresponds to the overlaps between separation boundaries learned from data generated from the original $\mathbf{W}$ distribution, and we provide an intuitive explanation illustrated in \textbf{Figure}~\ref{fig:pca_before}.
As discussed in~\cite{shen2020interfacegan}, for such strong entanglement, orthogonalization of the editing direction through projection does not work well, as it also removes the target direction. 
Similarly, the salient channels proposed in~\cite{chen2022exploring} for both attributes also overlap significantly, making channel filtering prone to fail to disentangle the attributes. 

\noindent \textbf{Disentangled edits with biased StyleGAN $\mathbf{W}$ directions.} Despite the biases in $\mathbf{W}$ space, previous works~\cite{harkonen2020ganspace,chen2022exploring} that learn editing directions based on $\mathbf{W}$ samples found a workaround to obtain disentangled edits. Instead of directly interpolating in $\mathbf{W}$, they apply the learned $\mathbf{W}$ directions to a subset of $\mathbf{W+}$ layers, which enables more localized controls. By limiting the changes to certain semantic regions, they successfully achieve disentanglement, e.g., only moving the mid-level $\mathbf{W+}$ layers to add lipstick and freezing the early layers that control global looks to avoid changing the gender. 

However, $\mathbf{W+}$ interpolation has limited capacity as the changes are mostly limited to specific semantic regions. When editing attributes that involve global-wise deformation, limiting the changes to specific $\mathbf{W+}$ layers sometimes results in the desired target effect not being present, an example of which is shown in \textbf{Figure}~\ref{fig:w+}. On the contrary, $\mathbf{W}$ space modifies the image on a global level with greater ranges of changes available. Nevertheless, $\mathbf{W+}$ interpolation is still useful as it provides us access to StyleGAN-generated images with minority attribute groups.  We hypothesize that, if these images can be reconstructed from latent codes in the $\mathbf{W}$ space, then such latent code cluster represents the low-density region needed for correcting the entangled $\mathbf{W}$ editing direction.
Therefore, we ask the following question: Instead of manipulating the learned editing directions, if we could obtain low-density samples, e.g., old people without eyeglasses in the $\mathbf{W}$ space, and create a less biased (more balanced) training distribution for the editing direction, would the newly trained direction be more disentangled? An intuition behind our hypothesis is shown in \textbf{Figure}~\ref{fig:pca} where we aim to create a more balanced distribution for age clusters in terms of samples wearing eyeglasses. 


\subsection{Learning Disentangled $\mathbf{W}$ Directions}
In order to debias the learned latent space distribution, efforts are needed to first identify the low-density regions and then acquire or generate the corresponding images. To automate this process and enable a large-scale correction, we introduce our method called SC$^2$GAN, which corrects the bias in the $\mathbf{W}$ distribution via self-corrected latent code samples. Given an entangled editing direction in $\mathbf{W}$, our propose to first interpolate $\mathbf{W}$ codes in $\mathbf{W+}$, which often shows more disentangled controls but does not generalize well for all attributes with the same hyper-parameters. Following such direction, we obtain edited images with localized changes corresponding to minority attribute groups, which will then be projected to the $\mathbf{W}$ space via GAN inversion. To enable disentangled editing, we re-train \cite{shen2020interfacegan, chen2022exploring} with this self-corrected latent distribution to learn the editing directions. We now describe the details of each step.

\noindent\textbf{Latent interpolation in $\mathbf{W}$ and $\mathbf{W+}$.} With an editing direction $f_a(\mathbf{w}) \subseteq \mathbb{R}^d$ learned, which could be a constant vector in the latent space, or a function of $\mathbf{w}$, interpolation in $\mathbf{W}$ space for 1 step with step size $s$ follows:
\begin{equation}
    \mathbf{w}^{\prime}=\mathbf{w}+f_a(\mathbf{w})*s
    \label{eq_interpolation}
\end{equation}
To interpolate in $\mathbf{W+}$, the editing process can be denoted as
\begin{equation}
\mathbf{w+}^{'} = E(f_a,\mathbf{w},\{i\},n,s)
\label{eq_w+}
\end{equation}, where the $i$th $\mathbf{W+}$ layers are edited following \textbf{Equation~\ref{eq_interpolation}} starting from $\mathbf{w}$ for $n$ steps with step size $s$. For example, $E(f_a,\mathbf{w},\{0,1,2,3\},3,0.5)$ means moving only the first 4 $\mathbf{W+}$ layers for 3 steps with step size 0.5, while $E(f_a,\mathbf{w},\{0..17\},3,0.5)$ is equivalent to interpolating in $\mathbf{W}$ space. As observed in multiple previous works~\cite{wu2021stylespace,harkonen2020ganspace,chen2022exploring}, compared to $\mathbf{W}$, the $\mathbf{W+}$ space from StyleGAN enables more localized controls, with $\mathbf{W}$ codes fed to layers at different resolutions controlling different abstraction levels, 
and the entanglement issue can be alleviated with spatial-wise editing in $\mathbf{W+}$. 

\noindent\textbf{Obtaining Self-corrected Samples in $\mathbf{W}$.} To verify our hypothesis above, we employ a latent optimization process and project the disentangled $\mathbf{W+}$ interpolation results to the $\mathbf{W}$ space following:
\begin{equation}
    \mathbf{w}^*= \underset{\mathbf{w}}{\arg \min } \ell(G(\mathbf{w}; \theta), G(\mathbf{w+}^{'} ; \theta))
    \label{eq_inv}
\end{equation} and we observe that the inverted $\mathbf{W}$ codes faithfully reconstruct the $\mathbf{W+}$ editing results and preserve the minority attribute groups well. In other words, these latent codes are self-corrected (disentangled) samples based on the original entangled editing directions, and they can be merged with the original $\mathbf{W}$ space samples to create a more balanced distribution for re-learning the editing directions. Assuming the original $\mathbf{W}$ space distribution has the entanglement issue between attribute $a_1$ and $a_2$, where the high-density regions mostly contain latent codes $\mathbf{W}$ with $({a_1}-,{a_2}+)$ and $({a_1}+,{a_2}-)$ semantics in the image space, hence changing the sign of $a_1$ by interpolating in $\mathbf{W}$ is likely to cause the opposite change in $a_2$. Through the process described above, we obtain $\mathbf{w}^*$ codes corresponding to images with $({a_1}+,{a_2}+)$ and $({a_1}-,{a_2}-)$ semantics in $\mathbf{W}$ space, hence the strength of $a_2$ in each $a_1$ cluster can be balanced by merging the original $\mathbf{w}$ codes with the self-corrected $\mathbf{w}^*$ codes. By retraining the editing direction for $a_1$ with the corrected distribution in $\mathbf{W}$, we decouple $a_1$ from the signs of $a_2$ and achieve disentangled and global controls.

\begin{figure*}[t!]
\begin{center}
\includegraphics[width=0.98\linewidth]{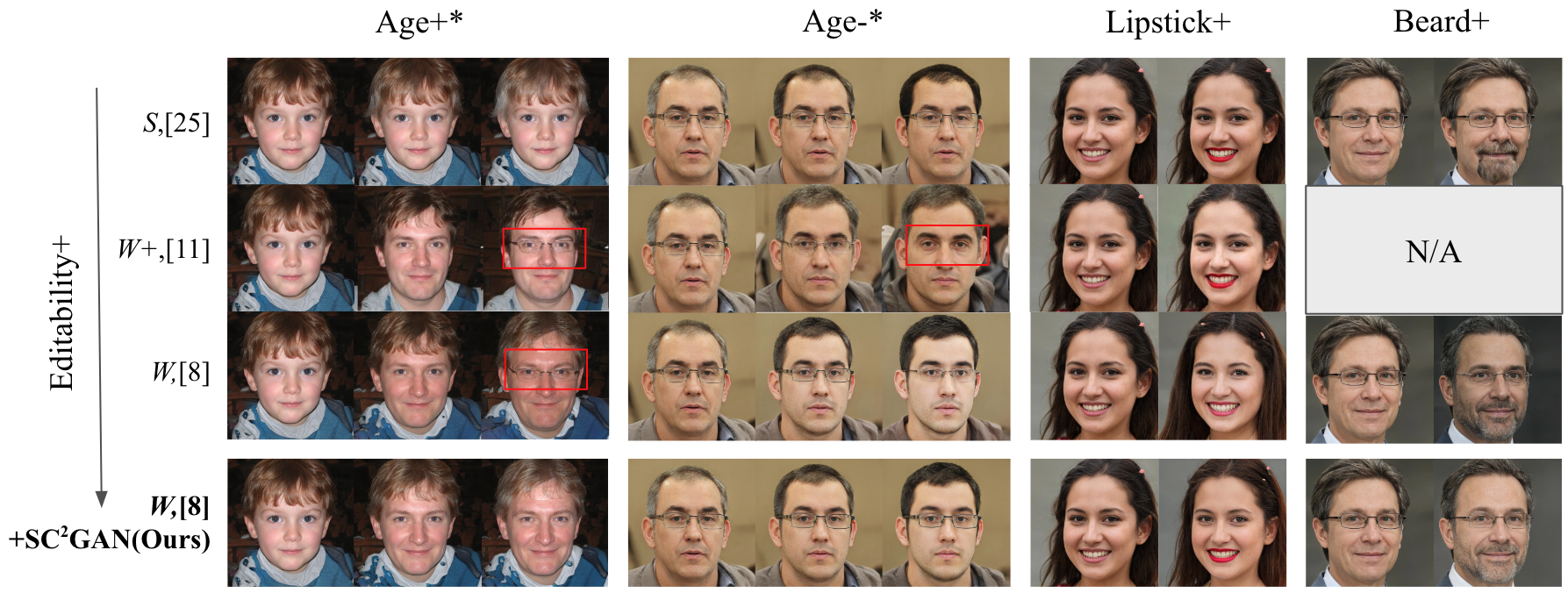}
\includegraphics[width=0.98\linewidth]{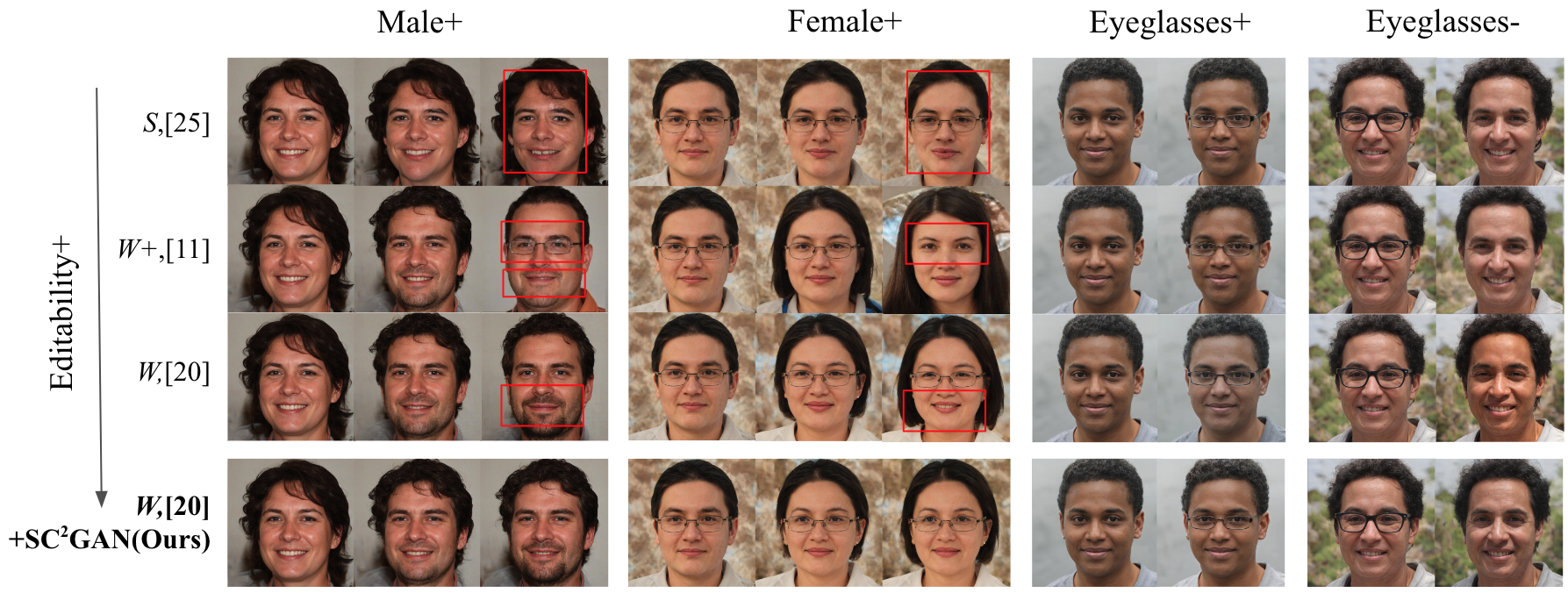}
\end{center}
\vspace{-0.2cm}
   \caption{Disentangled editing results. N/A means unsupported direction. *Age for~\cite{wu2021stylespace} is ``white hair'' and for the rest is ``age''. Within each group of images, left: source; middle, right: small and large interpolation direction. Our framework helps both methods~\cite{shen2020interfacegan,chen2022exploring} obtain more disentangled $\mathbf{W}$ controls and achieve better results than $\mathbf{W+}$~\cite{harkonen2020ganspace} and $S$ controls~\cite{wu2021stylespace} on global attributes like gender, and similar performance for localized ones like lipstick.}
\label{fig:ffhq_svm_ours}
\end{figure*}

\section{Experiments}
In this section, we apply our framework to existing supervised methods that learn editing directions based on $\mathbf{W}$ space latent code samples and obtain more disentangled directions. We first visualize such improvements by comparing the interpolation results before and after applying our framework, then quantify the amount of improvement for disentangling attribute pairs.

\subsection{Experiment Setup}
\noindent \textbf{Models.} We perform our experiments on the $\mathbf{W}$ space of StyleGAN2~\cite{karras2020analyzing} pretrained on FFHQ~\cite{karras2019style} with SVM-based~\cite{shen2020interfacegan} and gradient-based~\cite{chen2022exploring} editing directions. 
We sample 500k images and obtain pseudo labels for attributes gender, smile, eyeglasses, age, lipstick and beard with pre-trained attribute classifiers~\cite{karras2019style}.

\noindent \textbf{Learning the Original Directions.} Since our framework requires re-training of the learned editing directions, we first sample $\mathbf{W}$ latent codes corresponding to the images with the biggest\slash smallest logits from the classifier and follow~\cite{shen2020interfacegan} and~\cite{chen2022exploring} to learn the original $\mathbf{W}$ space editing directions.

\noindent \textbf{Learning the Disentangled Directions.} With the original editing directions learned by each method, we apply \textbf{Equation~\ref{eq_interpolation},~\ref{eq_w+},~\ref{eq_inv}} with the $\mathbf{W+}$ layer indices provided by ~\cite{harkonen2020ganspace,chen2022exploring} to the set of training samples to obtain the self-corrected samples. We re-train the directions from scratch using the merged dataset containing both self-corrected samples and the original $\mathbf{W}$ codes. More implementation details can be found in the Appendix.

\begin{figure*}[t!]
\begin{center}
\includegraphics[width=0.195\linewidth]{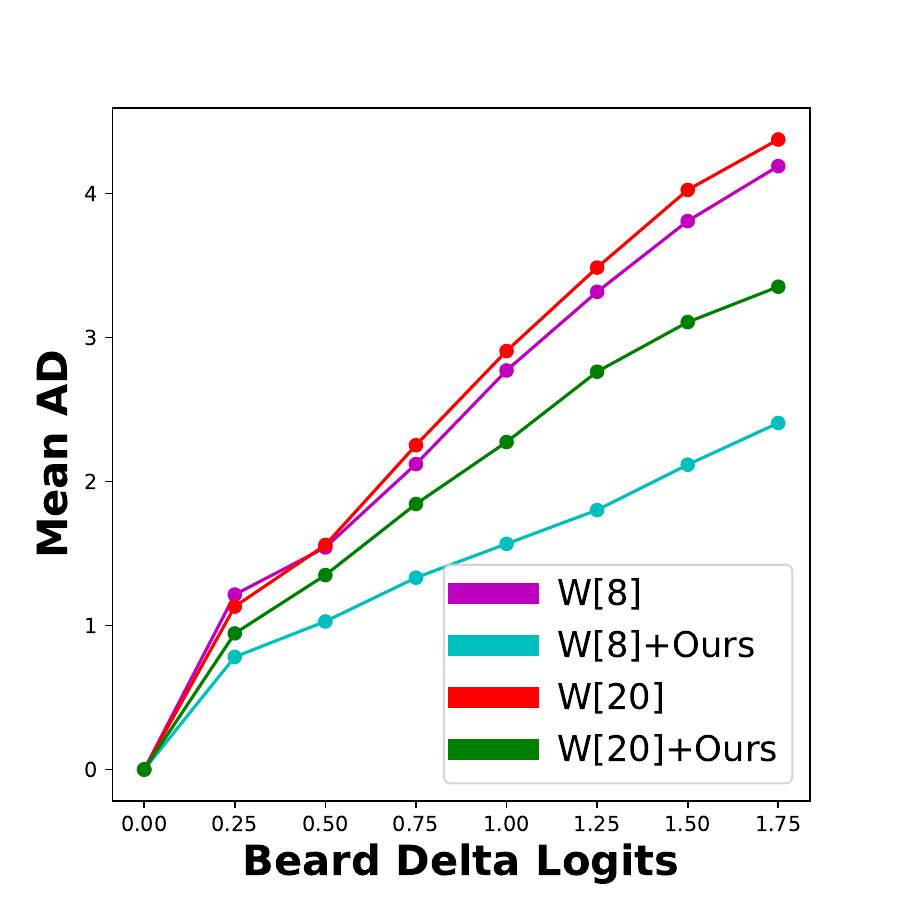}
\includegraphics[width=0.195\linewidth]{ 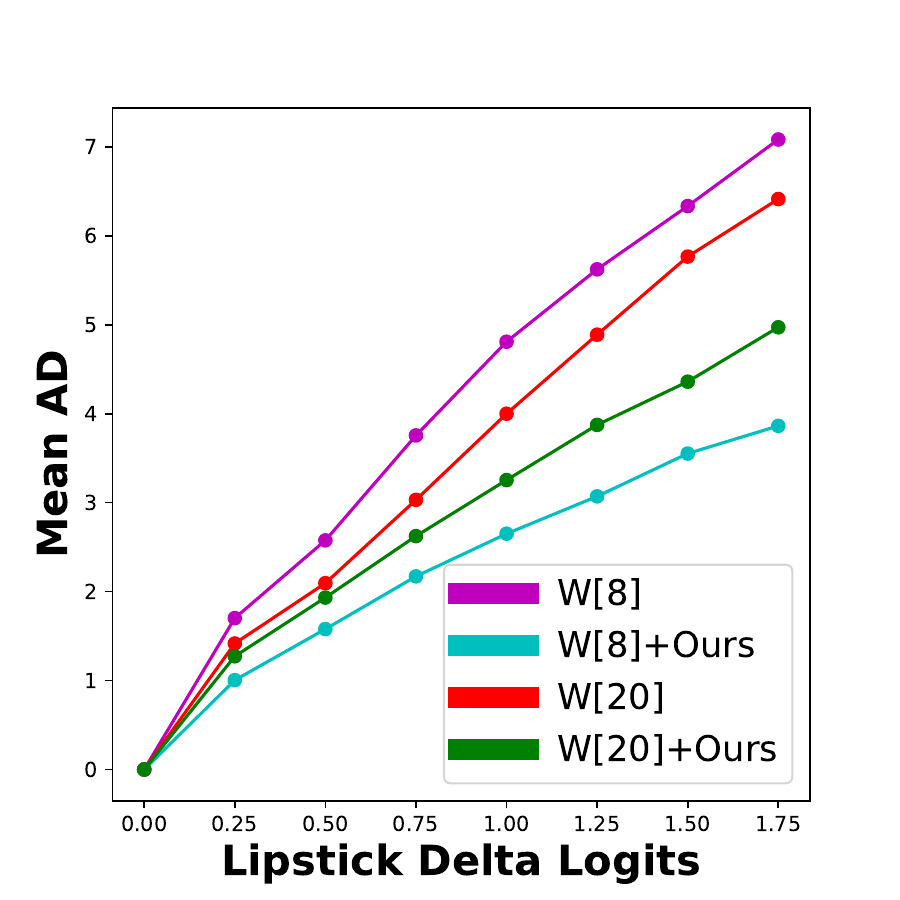}
\includegraphics[width=0.195\linewidth]{ 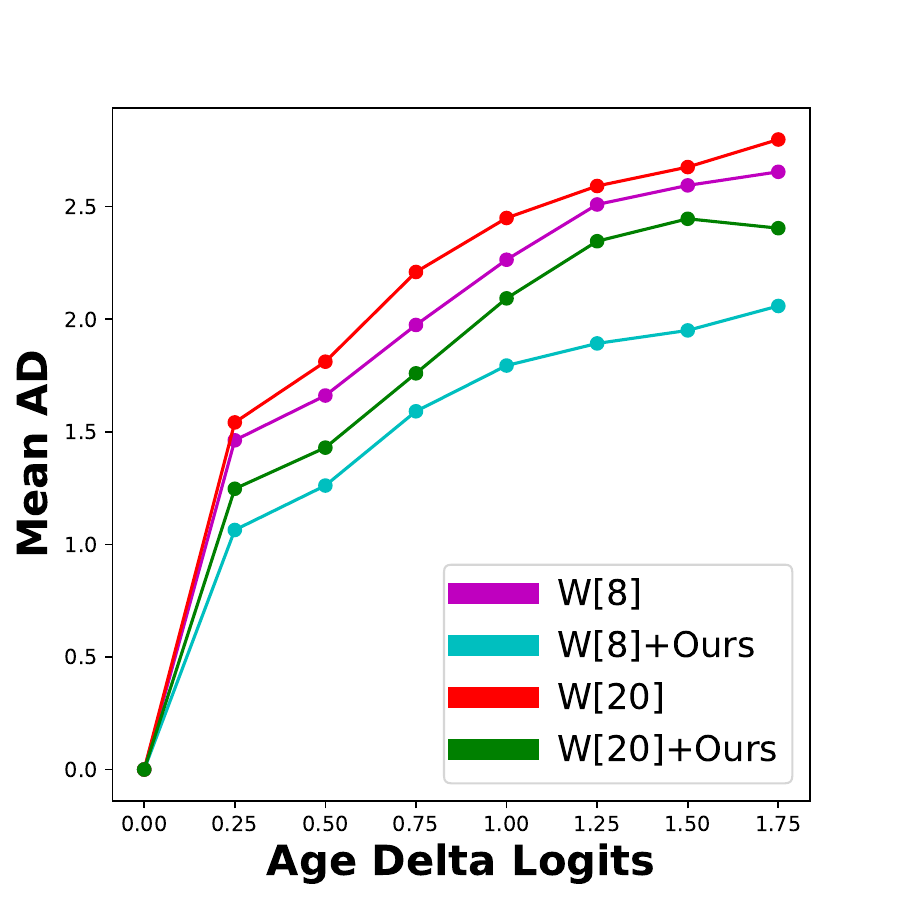}
\includegraphics[width=0.195\linewidth]{ 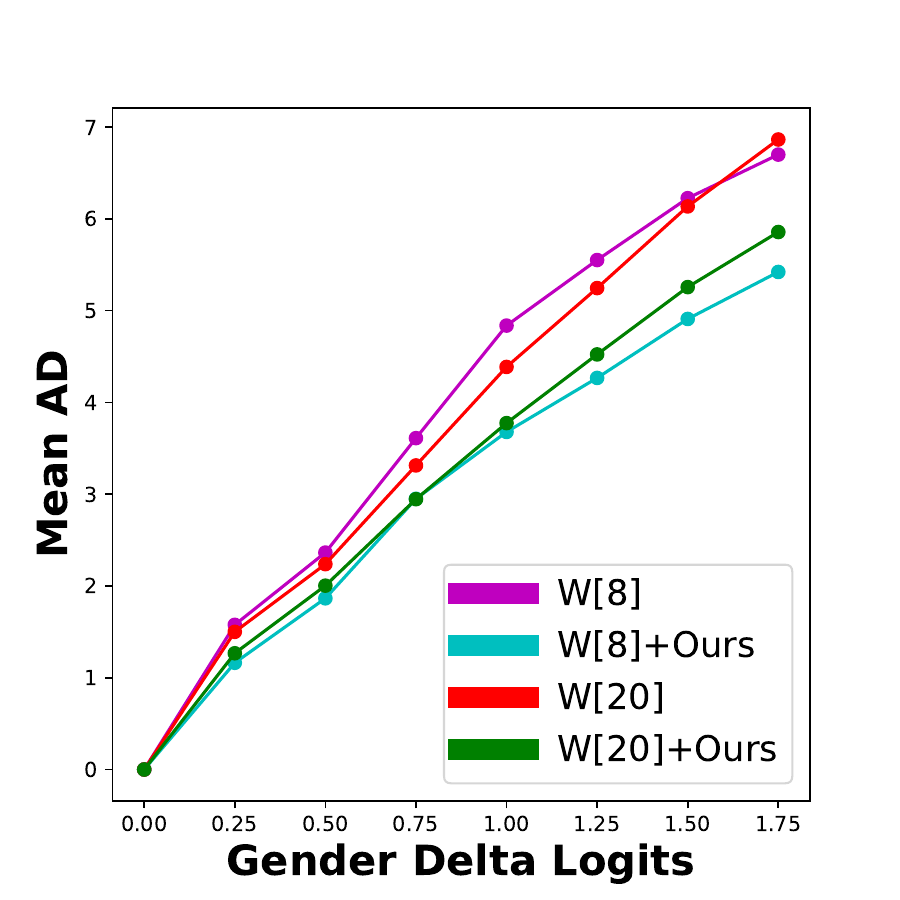}
\includegraphics[width=0.195\linewidth]{ 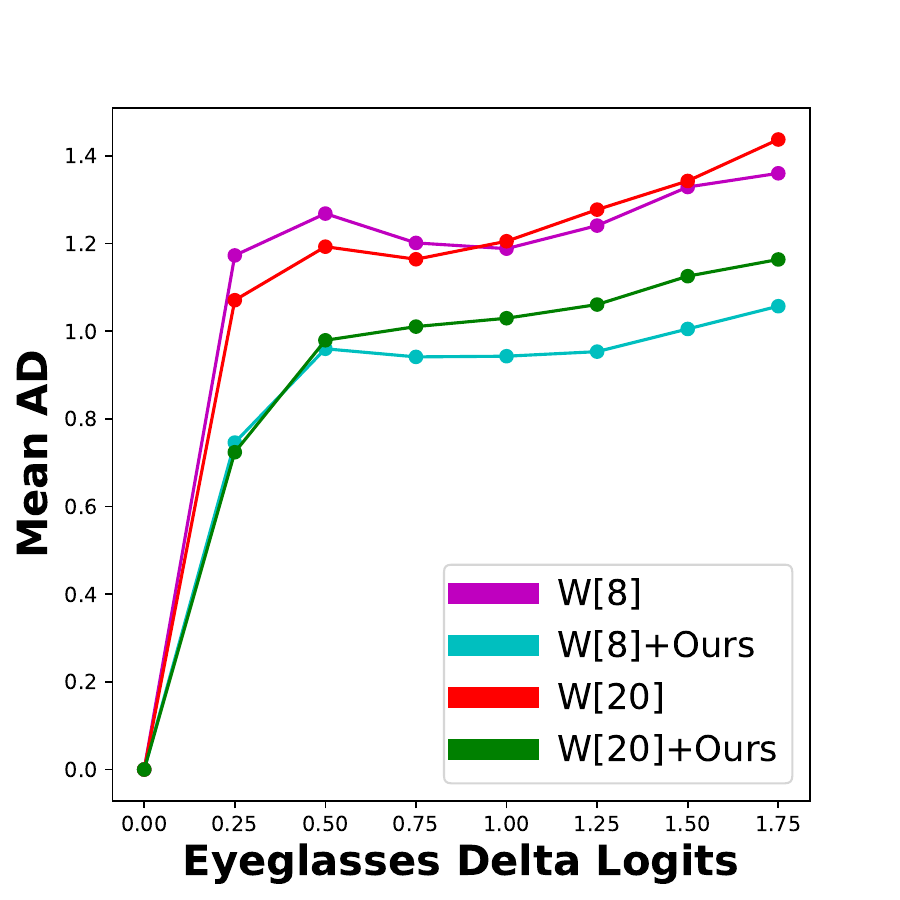}

\end{center}
   \caption{Attribute Dependency (AD), x-axis: normalized logit change in the target attribute, y-axis: mean of normalized logit changes in the others. Large y values mean strong entanglement. Our framework significantly reduces the amount of entanglement during interpolation in $\mathbf{W}$ space for both~\cite{shen2020interfacegan} and~\cite{chen2022exploring}.
   }
\label{fig:ad_main}
\end{figure*}
\begin{figure}[t]
\begin{center}
\includegraphics[width=\linewidth]{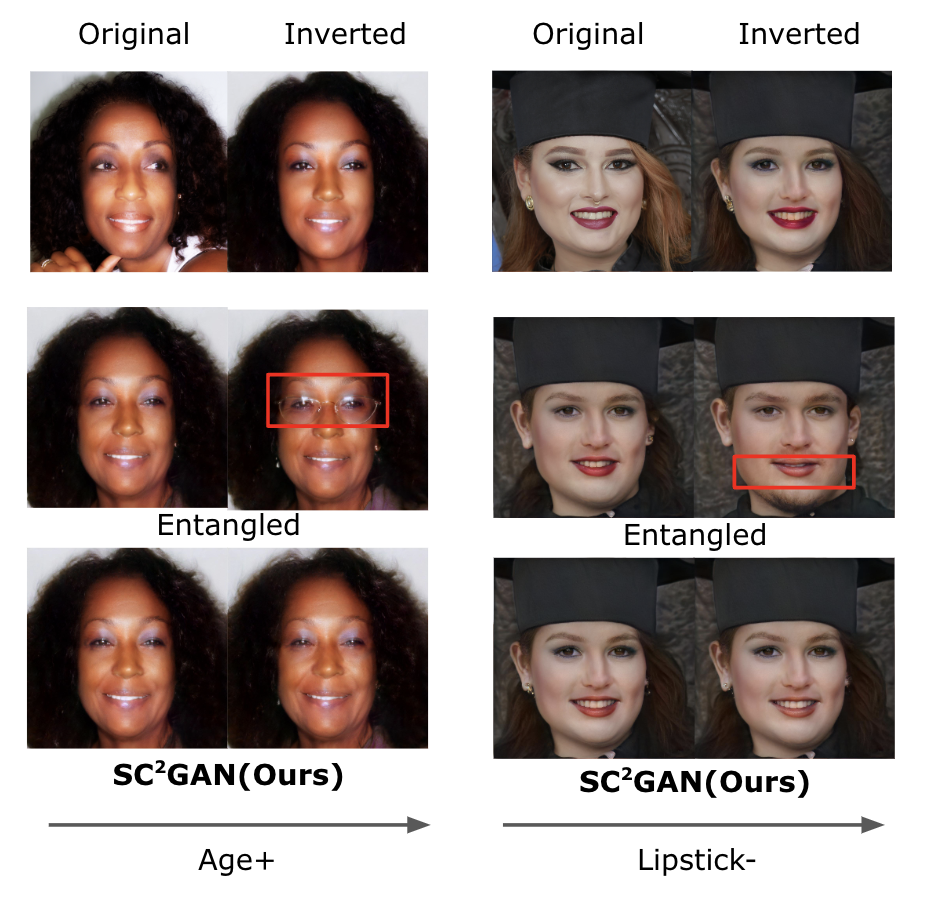}

\end{center}
   \caption{Disentangled controls with and without our framework applied to~\cite{shen2020interfacegan} for editing real images.}
\label{fig:svm_real}
\vspace{-0.45cm}

\end{figure}

\subsection{Disentangled Attribute Manipulation}
We first present qualitative results for attribute manipulation for gender, age, eyeglasses, lipstick and beard in \textbf{Figure}~\ref{fig:ffhq_svm_ours}, where we compare the original directions learned by Grad-Control~\cite{chen2022exploring} and InterFaceGAN~\cite{shen2020interfacegan} and editing directions after applying our framework to both methods. We also compare our results with methods to which our framework is not applicable. GANSpace~\cite{harkonen2020ganspace} learns meaningful directions in $\mathbf{W}$ by applying PCA to generator features and requires manual examination for semantic meanings, while StyleSpace~\cite{wu2021stylespace} finds locally activated semantic channels in $S$ space, which is $\mathbf{W+}$ layers with affine transformations applied. For both global attributes(age and gender) and local attributes(lipstick, eyeglasses and beard), our framework boosts disentanglement for both InterFaceGAN \cite{shen2020interfacegan} and Grad-Control \cite{chen2022exploring}. For instance, we achieve disentangled aging effects without eyeglasses added and decouple female direction from smile. GANSpace and StyleSpace suffer little from the entanglement issue, but the amount of change they make for global attributes is extremely limited, e.g., StyleSpace fails to synthesize more female effects, and GANSpace lacks the ability to generate aging effects. In the meantime, for local attributes, with our framework applied, InterFaceGAN and Grad-Control achieve performance similar to GANSpace and StyleSpace, which operate in spaces of much higher dimensions.

\subsection{Quantitative Results: Entanglement Analysis}

We quantify the level of entanglement based on Attribute Dependency(AD) proposed by~\cite{wu2021stylespace}. To compute the level of entanglement for one attribute $a$ with an editing method $f_a$, we first sample latent codes corresponding to images around the decision boundaries for the corresponding attribute classifier~\cite{karras2019style}, and interpolate them with fixed step sizes $d$ for 9 steps. At each step $s$, we compute the absolute change in logits for the target $x=\Delta l_s^{a}$ and the sum of absolute logits changes in the rest of the attributes, divided by the population standard deviation of each attribute $y=\frac{1}{|A|-1} \Sigma_{i \in A\setminus a}\frac{\Delta l_s^{i}}{\sigma l^i}$, where $A$ stands for the set of all attributes. Finally, we group all points with respect to ($\frac{x}{\sigma l^a}$) into buckets of ${(0,0.25],(0.25,0.5], \dots, (1.75,2]}$, and plot the midpoint for each bucket as the final x-value, mean of $y$ values within each bucket as the final y-value. We append the full algorithm and more details in the Appendix. As shown in \textbf{Figure}~\ref{fig:ad_main}, with our framework applied, the disentanglement in $\mathbf{W}$ editing direction improves significantly for ~\cite{chen2022exploring,shen2020interfacegan} on all attributes.

\subsection{Real Image Manipulation}
\textbf{Figure}~\ref{fig:svm_real} show the edited results on real images where entanglement exists for correlated features. Due to limited space, we demonstrate the age and lipstick edits following~\cite{shen2020interfacegan} with and without applying SC$^2$GAN. Our proposed approach achieves disentanglement while preserving the identity better.

\section{Ablation Studies}
\begin{figure}[t]
\begin{center}
\includegraphics[width=\linewidth]{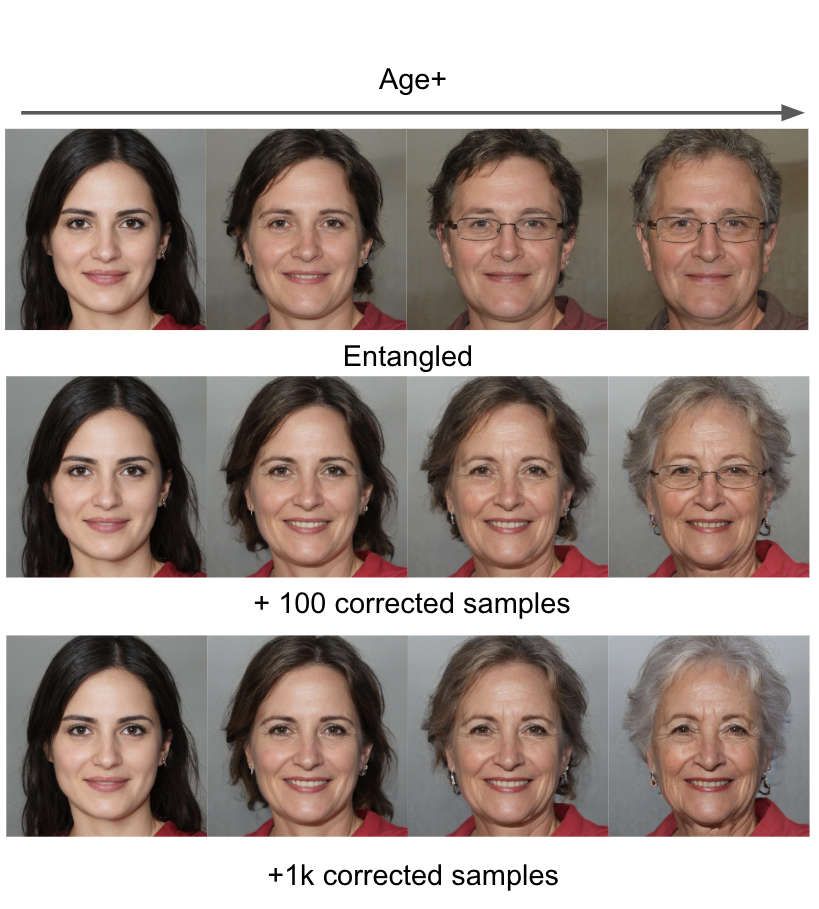}

\end{center}
   \vspace{-0.2cm}
   \caption{Comparison of different numbers of self-corrected samples added.}
\label{fig:moredata}
\end{figure}

\textbf{Number of Self-corrected Samples.} We qualitatively show how the number of self-corrected samples merged with the original $\mathbf{W}$ training data affects the overall editing directions learned by~\cite{chen2022exploring} in \textbf{Figure}~\ref{fig:moredata}, as their method can be trained only on a small dataset. With more self-corrected samples added, the original entanglement with eyeglasses is further minimized, with eyeglasses not appearing with similar aging effects present during interpolation. 




\begin{figure}[t]
\begin{center}

\vspace{-0.3cm}
\includegraphics[width=\linewidth]{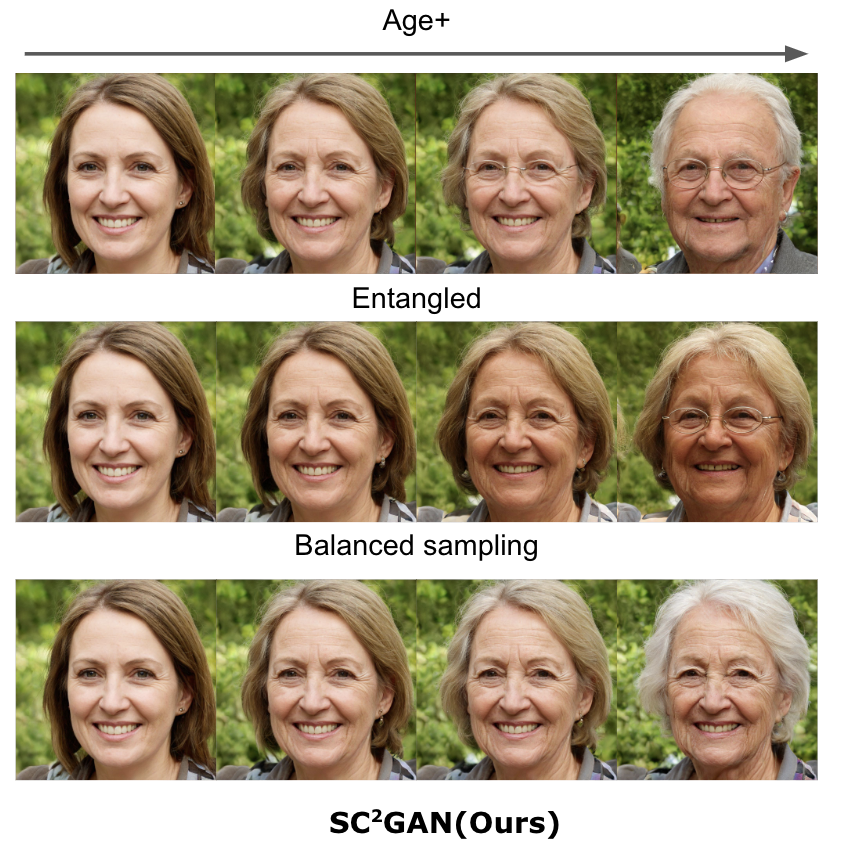}

\end{center}
\vspace{-0.2cm}
   \caption{Comparison of balanced sampling from the original $\mathbf{W}$ space and our framework. Balanced sampling-based direction makes the image unnaturally dark.}
   \vspace{-0.25cm}
\label{fig:balanced}
\end{figure}



\textbf{Directly Sampling Balanced Data.} An alternative approach that obtains the low-density area latent codes is to directly sample from such regions based on the pseudo labels of our image bank. However, as shown in \textbf{Figure}~\ref{fig:balanced}, although some entanglement can be alleviated with this approach, training with these samples could result in editing direction pointing to areas with lower image quality as the generator is not well-trained in those $\mathbf{W}$ regions. Furthermore, the amount of low-density data available for sampling is extremely limited, and as we take whatever is available given the scarcity of such data, these samples could lie close to the original separation boundary. Consequently, they may fail to provide a strong enough signal for the separation boundary to shift significantly.

\begin{figure}[t]
\begin{center}
\includegraphics[width=\linewidth]{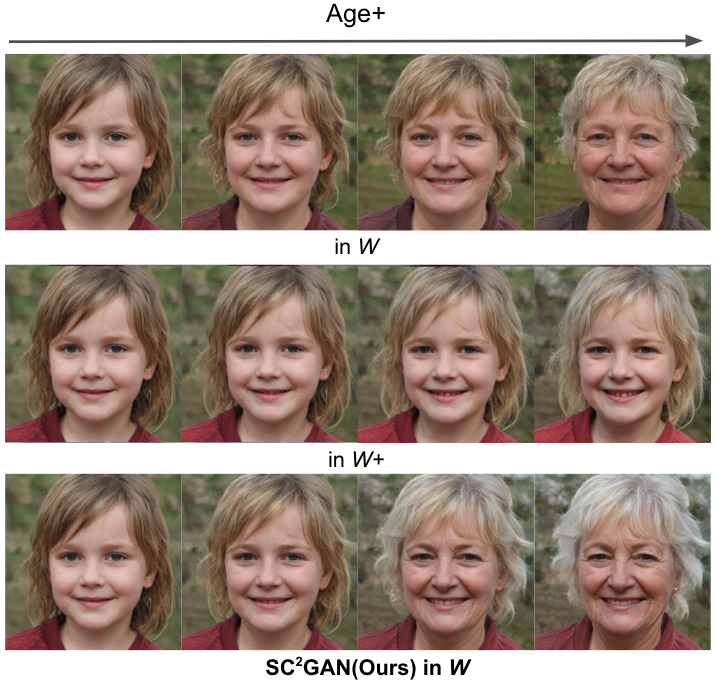}

\end{center}
   \caption{The difference between spatial-wise $\mathbf{W+}$ interpolation and $\mathbf{W}$ interpolation for increasing a child's age.}
\label{fig:w+}
\vspace{-0.15cm}

\end{figure}

\noindent \textbf{Comparison with $\mathbf{W+}$ Space Editing.} We base our work on findings of~\cite{karras2019style,harkonen2020ganspace,chen2022exploring} where $\mathbf{W+}$ space provides localized changes. Nevertheless, for attributes like aging, the editing involves great amounts of deformation of the original semantic regions, hence the localized $\mathbf{W+}$ space edits could fail to achieve the desired target effect, whereas interpolation in the $\mathbf{W}$ space is less prone to such failures as it modifies the image on a global level. We present an example in \textbf{Figure}~\ref{fig:w+}. Both directions learned with our framework applied to~\cite{chen2022exploring} and the original direction with $\mathbf{W+}$ interpolation do not suffer from the entanglement with eyeglasses. Yet, the latter fails to create aging effects like saggy cheeks and ptosis of eyelids, with the changes mostly limited to the initial semantic regions.

\section{Conclusion}
We study the entanglement problem in the $\mathbf{W}$ space of StyleGAN2 and propose SC$^2$GAN, a simple yet effective method that generates self-corrected samples in low-density regions to obtain disentangled controls.
 With these self-corrected samples added to the original $\mathbf{W}$ distribution, we learn decoupled separation boundaries that enable disentangled editing.
Overall, our framework shows strong capabilities to disentangle attributes with similar separation boundaries and salient channels in the original latent space, and works well in both local and global attribute manipulations.

\bibliographystyle{ieee_fullname}
\nocite{*}
\bibliography{egbib}

\end{document}